\documentclass{bvm2022} % Do NOT change this line

\makeatletter
\let\blx@rerun@biber\relax
\makeatother

\begin{document}

% Do not change anything in the preamble (anything above \begin{document}) except for the specification of the bibliography file, any additional changes will be lost

% use the \selectlanguage command to select the language in which your proceedings are written

%\selectlanguage{ngerman} % German
\selectlanguage{english} % English

% Indication of the title of your contribution
\title{Superpixel Pre-Segmentation of HER2 Slides for Efficient Annotation}
% If you write a short paper/abstract, the title must start with "Abstract:".
% \title{Abstract: Bildverarbeitung für die Medizin 2022}

% Optional specification of subtitle
%\subtitle{Guidelines for the Creation of the Print-ready Contributions}

% titlerunning appears in the header of every second page
% LaTeX generates this automatically from your contribution title
% However, if it is too long, the message "Title Suppressed Due to Excessive Length" appears instead.
% In this case, specify an abbreviated form of the title here
\titlerunning{Superpixel Pre-Segmentation of HER2 Slides}

% Please indicate all authors involved
% To allow us to correctly identify the last name of each author, indicate it using the \lname{} command.
% If more than one institute is involved, list the number of the institute(s) (see below) with \inst{} after the respective author. If only one institute is involved, omit this.
% Separate all authors with a comma
\author{
	Mathias \lname{Öttl} \inst{1}, 
	Jana \lname{Mönius} \inst{2}, 
	Christian \lname{Marzahl} \inst{1}, 
	Matthias \lname{Rübner} \inst{4}, 
	Carol~I. \lname{Geppert} \inst{2}, 
	Arndt \lname{Hartmann} \inst{2},
	Matthias~W. \lname{Beckmann} \inst{4},
	Peter \lname{Fasching} \inst{4},
	Andreas \lname{Maier} \inst{1},
	Ramona \lname{Erber} \inst{2},
	Katharina \lname{Breininger} \inst{3}
}

%\author{Mathias Öttl$^1$, Jana Mönius$^2$, Christian Marzahl$^1$, Matthias Rübner$^4$, Carol I. Geppert$^2$, Arndt Hartmann$^2$, Matthias W. Beckmann$^4$, Peter A. Fasching$^4$, Andreas Maier$^1$, Ramona Erber$^2$, Katharina Breininger$^3$}

% Enter the authors here as you want them to appear in the header
% Name only the surnames
% Depending on the number of authors involved, follow the examples below
% \authorrunning{Meier} - one author
% \authorrunning{Meier \& Müller} - two authors
% \authorrunning{Meier, Müller \& Schulze} - three authors
% \authorrunning{Meier et al.} - more than three authors
\authorrunning{Öttl et al.}

% Specify the institutes involved
% In case of participation of more than one institute, each institute shall be preceded by an ascending number with \inst{}.
% If only one institute is involved, omit the corresponding number.
% Separate individual institutes with \\
\institute{
\inst{1} Pattern Recognition Lab, Friedrich-Alexander-Universität Erlangen-Nürnberg (FAU)\\
\inst{2} Institute of Pathology, University Hospital Erlangen, Friedrich-Alexander-Universität Erlangen-Nürnberg (FAU), Comprehensive Cancer Center Erlangen-EMN (CCC ER-EMN)\\
\inst{3} Department Artificial Intelligence in Biomedical Engineering, Friedrich-Alexander-Universität Erlangen-Nürnberg (FAU)\\
\inst{4} Department of Gynecology and Obstetrics, University Hospital Erlangen, Friedrich-Alexander-Universität Erlangen-Nürnberg (FAU)
}

% Enter the e-mail address of the corresponding author
\email{mathias.oettl@fau.de}

\maketitle

% Abstract of your contribution, only for long contributions
% Do NOT use \begin{abstract} ... \end{abstract} for short articles
\begin{abstract}
Supervised deep learning has shown state-of-the-art performance for medical image segmentation across different applications, including histopathology and cancer research; however, the manual annotation of such data is extremely laborious. In this work, we explore the use of superpixel approaches to compute a pre-segmentation of HER2 stained images for breast cancer diagnosis that facilitates faster manual annotation and correction in a second step. Four methods are compared: Standard Simple Linear Iterative Clustering (SLIC) as a baseline, a domain adapted SLIC, and superpixels based on feature embeddings of a pretrained ResNet-50 and a denoising autoencoder. To tackle oversegmentation, we propose to hierarchically merge superpixels, based on their content in the respective feature space. When evaluating the approaches on fully manually annotated images, we observe that the autoencoder-based superpixels achieve a 23\% increase in boundary F1 score compared to the baseline SLIC superpixels. Furthermore, the boundary F1 score increases by 73\% when hierarchical clustering is applied on the adapted SLIC and the autoencoder-based superpixels. These evaluations show encouraging first results for a pre-segmentation for efficient manual refinement without the need for an initial set of annotated training data.

\end{abstract}

\section{Introduction} \label{3240-intro}
Breast cancer, a common type of cancer that mainly affects women, can be classified into different subtypes that show differences in their aggressiveness, treatment response and prognosis~\cite{3240-00}. Human Epidermal growth factor Receptor 2 (HER2) positive breast cancer is associated with a poor prognosis but targeted therapies exist that can reduce the mortality of patients suffering from this subtype~\cite{3240-00}. 
To determine whether a specific treatment may benefit a patient, HER2 expression is visualized using a specific staining and scored by pathologists~\cite{3240-00} into four classes (0, 1+, 2+, or 3+): Tumors with a score of 0 or 1+ are considered HER2-negative, and cases with a score of 3+ are regarded as HER2 positive. In routine diagnostics, breast cancer samples with a 2+ staining are further analyzed 
%for gene copy number (GCN) alteration of HER2 using in situ hybridization (ISH) 
to determine positive or negative HER2 status.
Recently, machine learning has been incorporated into HER2 research to better understand growth and proliferation and support the analysis of this data.
For example, deep learning methods were used to segment the cell membranes~\cite{3240-01}. These intermediate results were then utilized to determine the HER2 score.
Label generation for these methods is tedious and time-consuming since fine-grained annotations are required. In this work, we evaluate superpixel methods under the premise of reducing HER2 annotation effort and to ease future work on comprehensible deep learning methods~\cite{3240-02}.
Single tumor cells per slide may present with different HER2 staining intensities and HER2 expression patterns, however, local clusters of cells often share the same HER2 score ~\cite{3240-03}. The size of areas with homogeneous HER2 score can vary greatly, and these areas show complex staining textures which motivates the use of superpixels in this context. As a baseline, we use SLIC superpixels, which are reasonably robust when the data has a non-uniform texture and achieve good results compared to other superpixel methods like the Felzenszwalb algorithm, quick shift or the normalized cuts algorithm~\cite{3240-04}. Furthermore, we investigate different variants for this specific application, namely SLIC adapted to the color properties of HER2 stained images, as well as superpixels based on the latent representation of pretrained networks and denoising autoencoders~\cite{3240-05}. Due to the varying size of areas with homogeneous HER2 expression, the constant size of SLIC-based superpixels would result in strong over- or undersegmentation depending on the selected superpixel size. To counteract this effect, we incorporate an additional hierarchical clustering stage to merge superpixels with similar content and evaluate the effect on the resulting pre-segmentation as a further contribution of this work.

\section{Materials and methods}
In this section, we first describe the four methods used to calculate superpixels, followed by the hierarchical clustering approach applied to all four superpixel methods. Furthermore, we present the data used for training and evaluation as well as the evaluation metrics.

\subsection{Superpixel methods}

As \textit{baseline}, we use the standard SLIC algorithm described by Achanta et al.~\cite{3240-04}. Furthermore, we adapt the representation of the HER2 images as follows:

\textit{Adjusted SLIC:} As a preprocessing step, images are converted from RBG to Haematoxylin-Eosin-DAB (HED) color space~\cite{3240-06} to utilize prior knowledge about the color distribution of HER2 stained images. Originally developed for Hematoxylin and Eosin (H\&E) stained images, we use it to obtain a more uniform distribution of the hematoxylin color channel by filtering the respective image channel with a large Gaussian kernel. This channel mainly depicts cell nuclei which are not of interest for this application. Accordingly, the values of this channel are reduced to force the subsequently applied SLIC algorithm to focus on the color channels of the HER2 stain and less on the cell nuclei.

\textit{Pre-trained ResNet-50:} A ResNet-50, pretrained as feature extractor for a mask R-CNN on the COCO segmentation dataset, is utilized to compute lower resolution features vectors from the fourth stage of the network. These feature vectors, called embeddings, are upsampled to the original image resolution and serve as input for the SLIC algorithm.

\textit{HER2 autoencoder:} To obtain embeddings that are targeted to HER2 data, we train a denoising autoencoder to reconstuct augmented input images with the goal to determine a robust internal data representation. Our autoencoder has three levels, each with one convolutional layer in the encoder and one transposed convolutional layer in the decoder. ReLU is used as activation function. The network is trained using Stochastic Gradient Descent (SDG) with momentum (learning rate 0.01, momentum 0.9), batchsize of 32 and mean squared error loss function, until convergence of loss on the validation set. At lowest level, 64-dimensional embeddings are calculated, which are also upsampled to the original image resolution and serve as input for the SLIC algorithm.

\subsection{Hierarchical clustering}

To address the problem of over- or undersegmentation that occurs when SLIC is used to segment areas of vastly different sizes, hierarchical clustering is applied. The benefits of clustering superpixels into larger regions were shown in~\cite{3240-07}. In this work, superpixels are described by the mean value of their pixels, which can either be a color or a feature vector. Based on this mean value, the superpixels are combined using agglomerative clustering. As clustering criterion ward~\cite{3240-08} is used, which enforces the variance within clusters to be low. In addition, only neighboring superpixels are allowed to be merged.

\subsection{Dataset}

The dataset used in this work consists of image regions from 20 tissue sections (five for each HER2 score), originating from 20 different patients. These sections were immunohistochemically stained for HER2 and scanned using a PANNORAMIC 1000 scanner from 3DHistech, using a 20x objective, and are available as digital images. A total of 84 images patches of size 1.5\,mm$\times$1.5\,mm were extracted to serve as the dataset. The patches were randomly split 60/20/4 on slide level into a training, validation and test set for autoencoder training and evaluation, with equal distribution of HER2 score across all subsets. The four test patches for evaluation were manually annotated into HER2 scored tumor areas, non-tumor areas and staining artefacts.
These areas were segmented using polygons in the EXACT tool~\cite{3240-09}. Annotations were performed by a medical student and reviewed by a board-certified pathologist.

\subsection{Evaluation metrics}

The evaluation metrics used in this work are described in detail in~\cite{3240-05}.
\textit{Achievable segmentation accuracy (ASA} describes the segmentation coverage of the superpixels and indicates whether superpixels overlap multiple ground truth segments.
\textit{Boundary F1 score (F1)} is a measure for the quality of the boundary segmentation. The score is a equal weighted combination of boundary recall and boundary precision.
Since the exact position of a boundary is often fuzzy, a tolerance is used in boundary measures. HER2 data has high resolution and tissue areas are hard to precisely delineate from each other, therefore a tolerance of 15 pixels (3.75\,\textmu m) is used for the evaluation of the different approaches.

\section{Results}

\begin{figure}[b]
	\centering
	\setlength{\figwidth}{0.49\textwidth}
	\begin{subfigure}{\figwidth}
		\includegraphics[width=\textwidth]{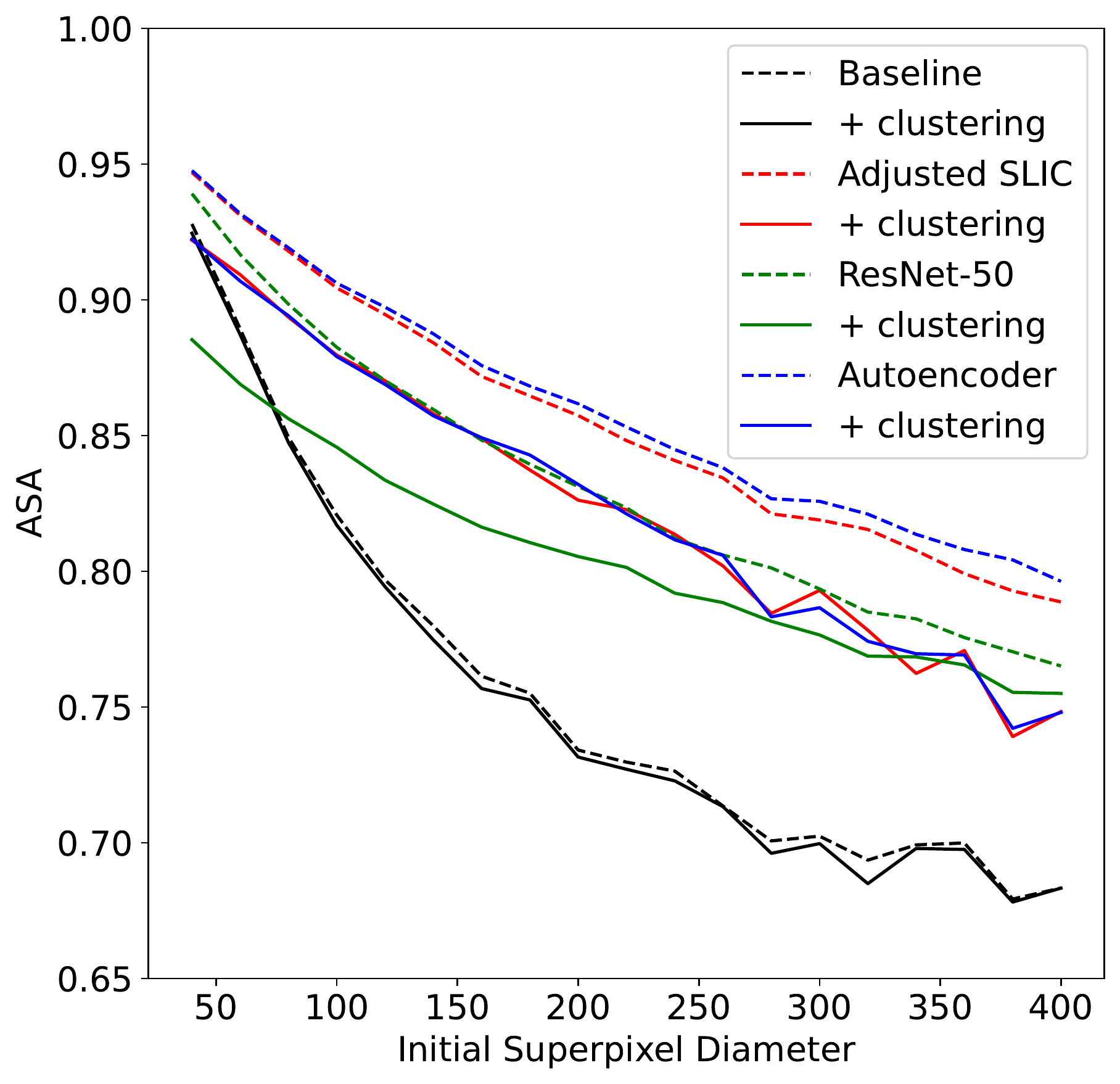}
		\caption{ASA metric}
	\end{subfigure}
	\hfill
	\begin{subfigure}{\figwidth}
		\includegraphics[width=\textwidth]{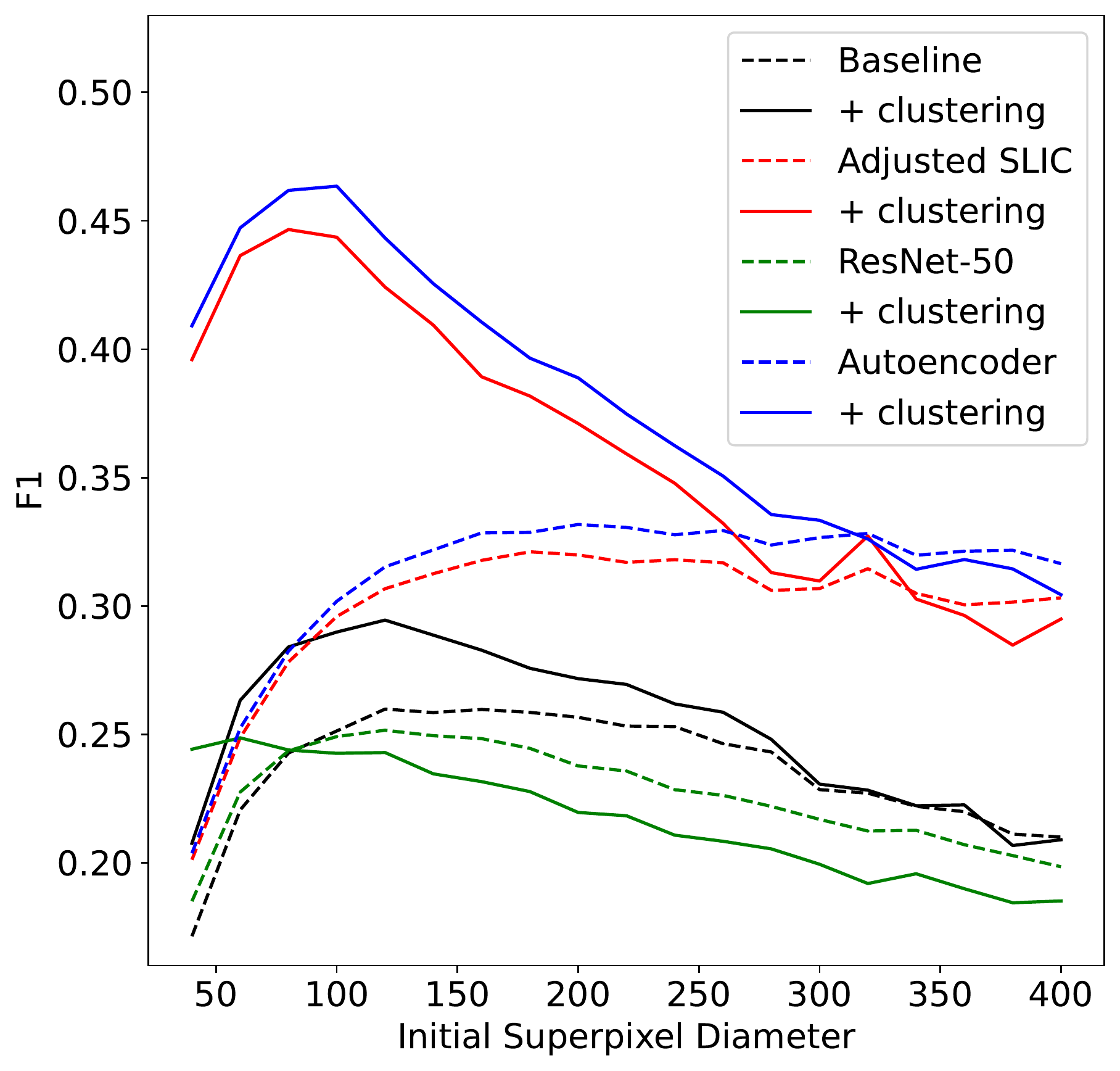}
		\caption{F1 score}
	\end{subfigure}
	\caption{Plots of achievable segmentation accuracy (ASA) (a) and boundary F1 score (b) for the different methods for different superpixel sizes.}
	\label{3240-fig-01}
\end{figure}

\begin{figure}[b]
    \centering
	\setlength{\figwidth}{0.31\textwidth}
	\begin{subfigure}{\figwidth}
		\includegraphics[width=\textwidth]{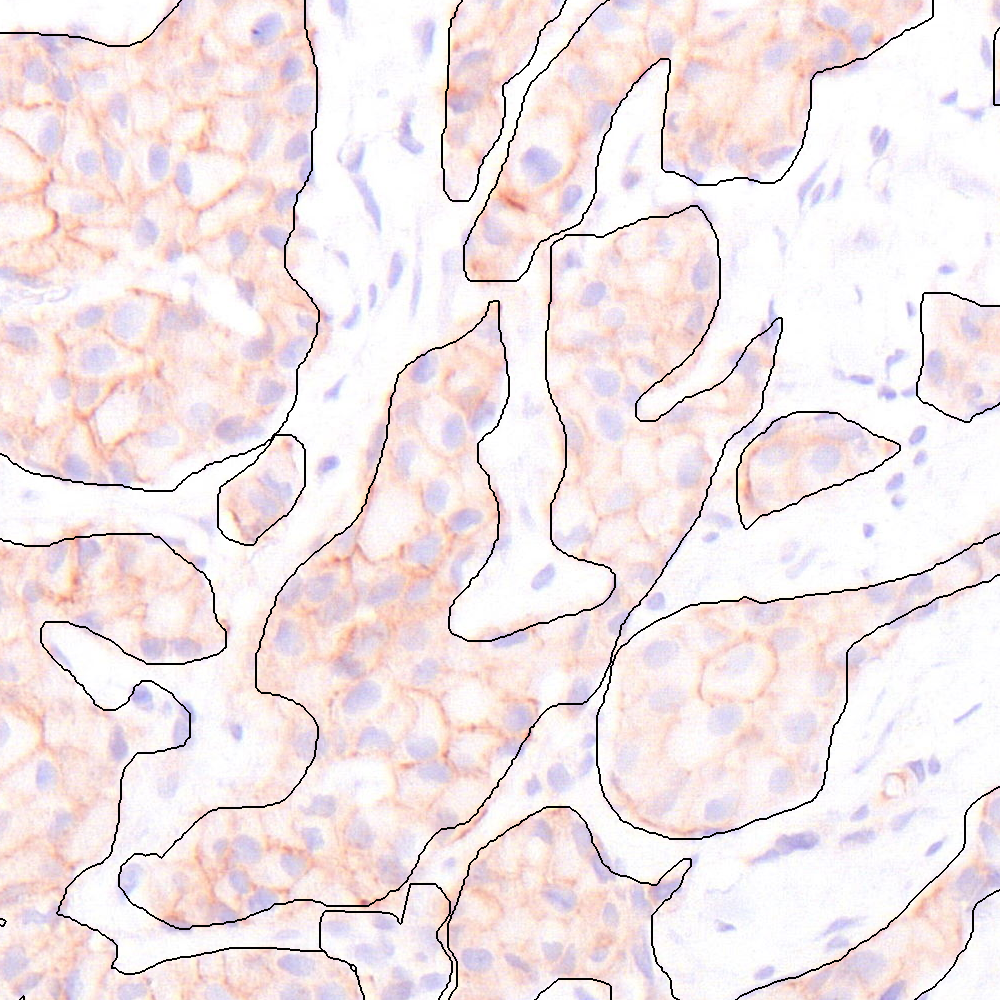}
		\caption{Ground truth}
	\end{subfigure}
	\quad
	\begin{subfigure}{\figwidth}
		\includegraphics[width=\textwidth]{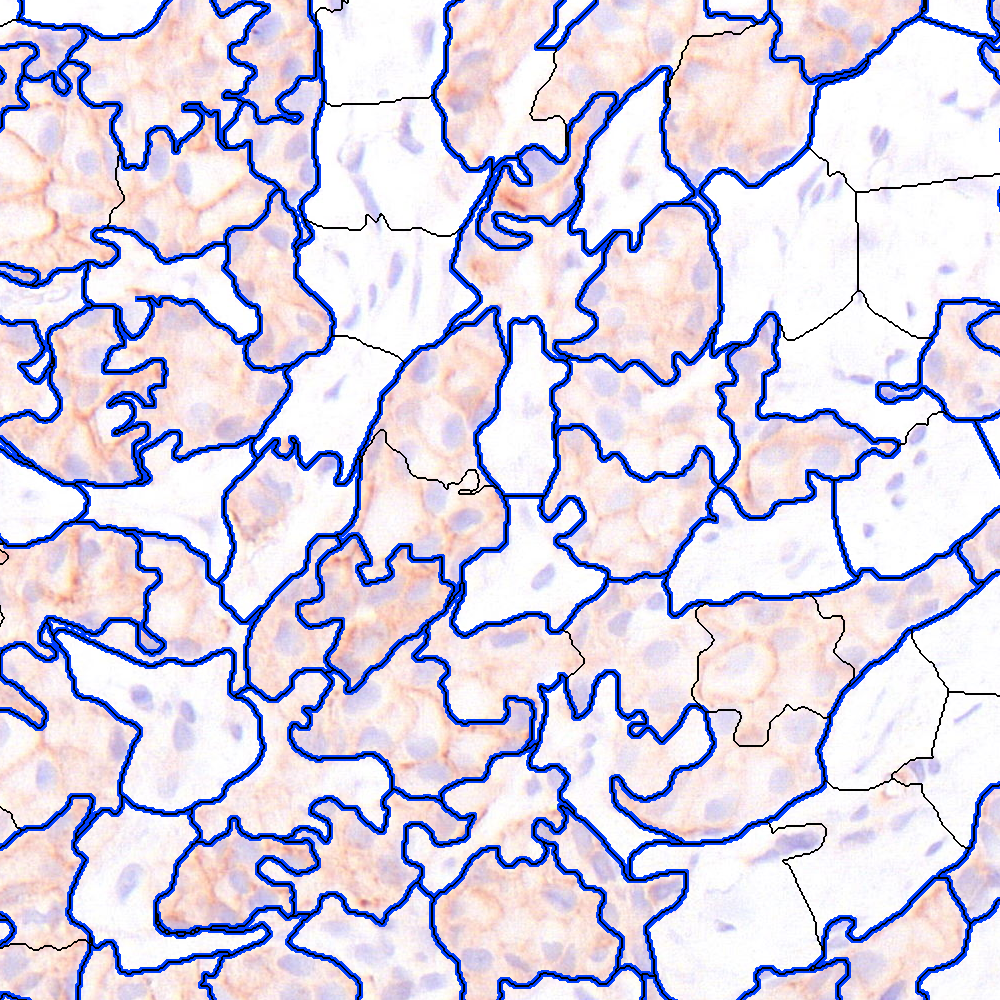}
		\caption{SLIC}
	\end{subfigure}
	\quad
	\begin{subfigure}{\figwidth}
		\includegraphics[width=\textwidth]{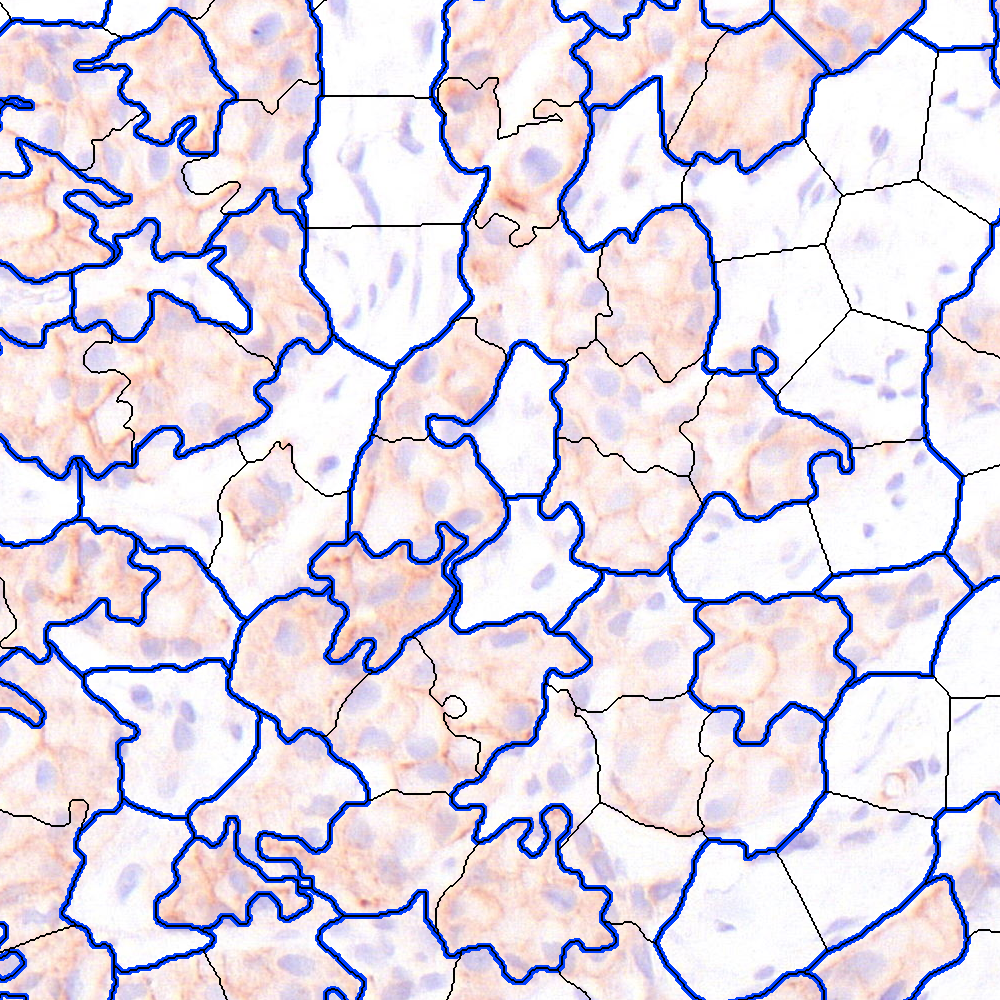}
		\caption{Adjusted SLIC}
	\end{subfigure}
	\quad
	\begin{subfigure}{\figwidth}
		\includegraphics[width=\textwidth]{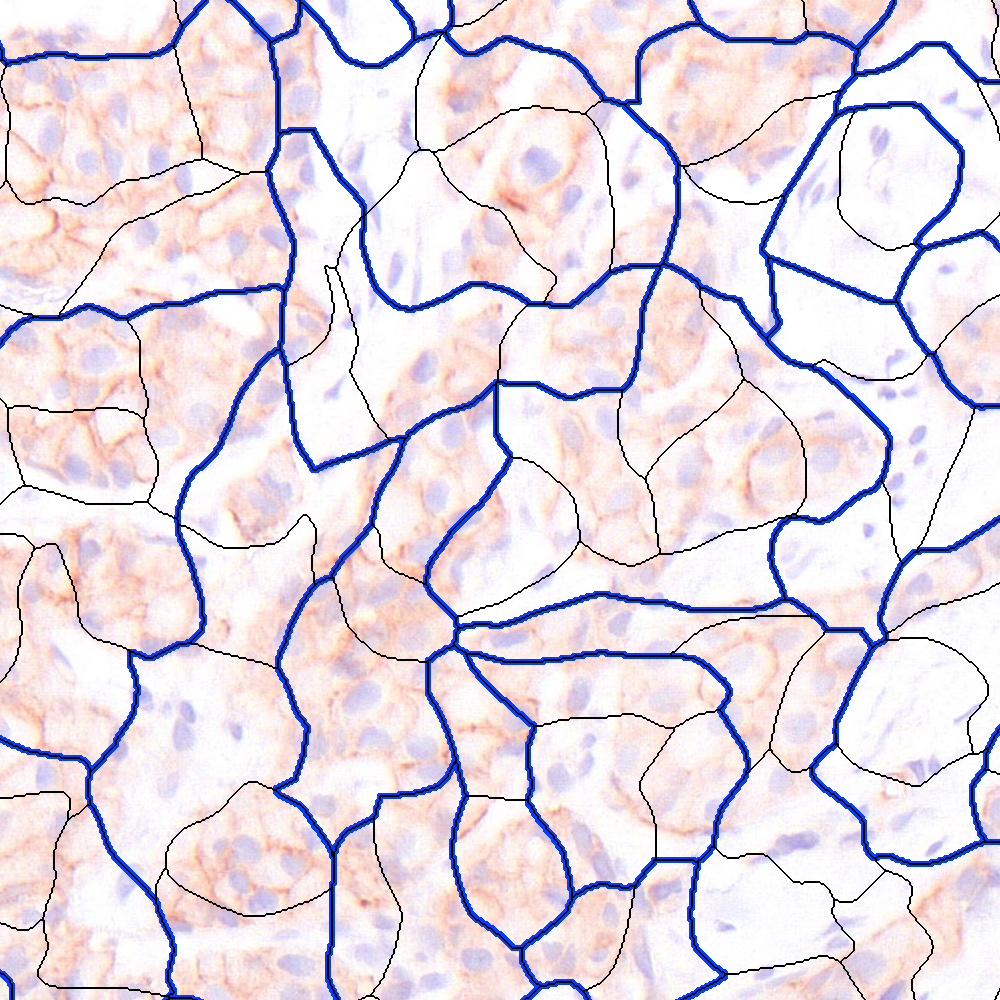}
		\caption{ResNet-50}
	\end{subfigure}
	\quad
	\begin{subfigure}{\figwidth}
		\includegraphics[width=\textwidth]{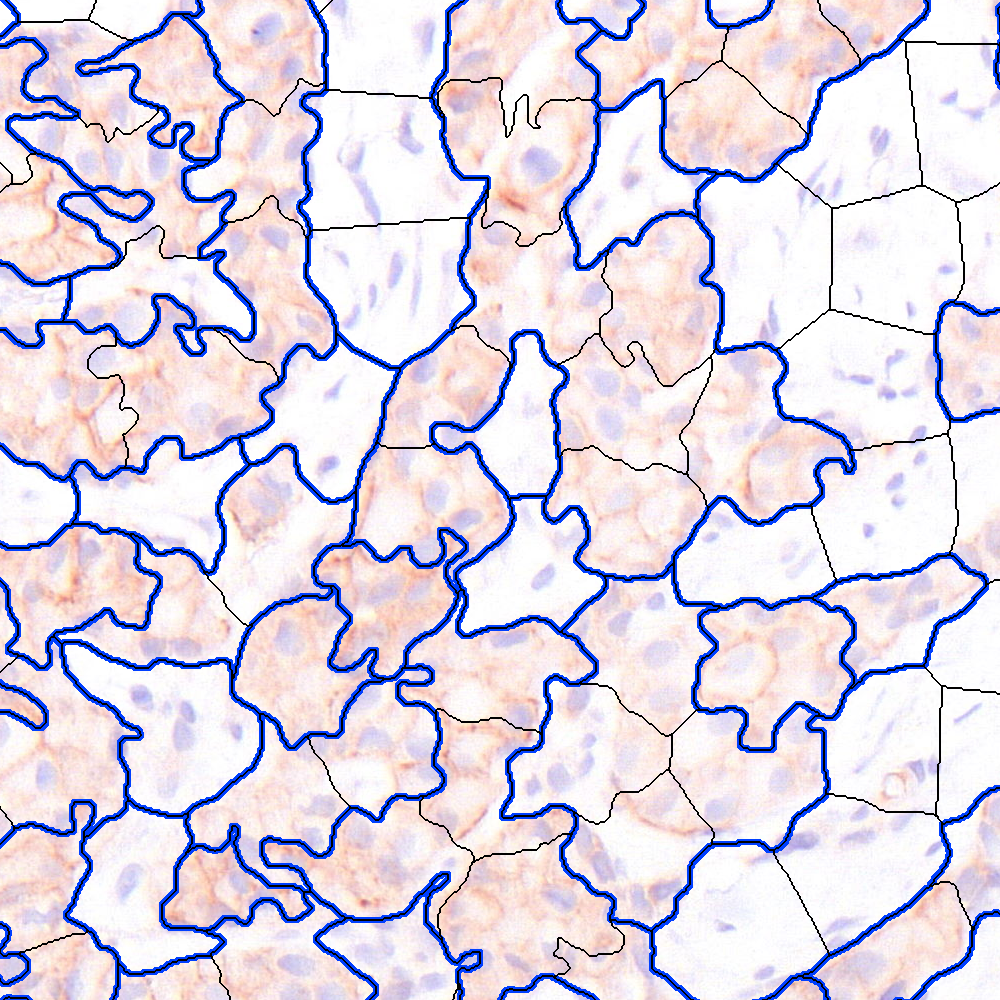}
		\caption{Autoencoder}
	\end{subfigure}
	\caption{Results for the superpixel methods at a inital superpixel diameter of 120 pixels. The initial superpixels are drawn with black outlining, while the clustered superpixels are visualized in blue.}
	\label{3240-fig-02}
\end{figure} 

Figure~\ref{3240-fig-01} shows the metrics for the superpixel methods with and without hierarchical clustering. As expected, the ASA score for all versions is close to one for the initially selected superpixel size and decreases with higher superpixel diameter. A stronger decrease is visible for the baseline method, which indicates that the baseline method computes superpixels that wrongly cross tissue boundaries. For all methods the ASA score is slightly lower when hierarchical clustering is utilized, showing that some incorrect superpixel merges occur.
With respect to the boundary F1 score, the baseline (peak value: 0.26) is already outperformed by adjusted SLIC (peak value: 0.32) and the autoencoder (peak value: 0.33) when no hierarchical clustering is applied. These values represent an improvement of 23\% and 27\% respectively. With hierarchical clustering, adjusted SLIC and the autoencoder both achieve a peak value of 0.45, representing an improvement of 73\% compared to the baseline without hierarchical clustering. The pretrained ResNet-50 underperforms the baseline in all cases, except for small initial superpixel diameters.
Figure~\ref{3240-fig-02} shows example regions of the ground truth and the output of the superpixel methods with (blue) and without hierarchical clustering (black). For all approaches the visual results are in line with the reported metrics.

\section{Discussion}

In this work, several superpixel methods and an additional hierarchical clustering step were evaluated.
We showed that superpixels from a domain adjusted SLIC and from a HER2 autoencoder outperformed a baseline SLIC in terms of boundary F1 score and achievable segmentation accuracy. These results highlight that the HER2 representations used in this methods are superior for calculating SLIC superpixels and should be considered in future segmentation task on HER2 data.
Additionally, we showed that hierarchical clustering of superpixels achieves significantly higher boundary F1 scores, while decreasing the ASA score slightly. This technique reduces the annotation effort of HER2 data even further and will help to speed up the creation of ground truth annotations.
Although these results are promising, the following limitations should be noted. The test dataset consists of one labelled image for each of the four HER2 scores, which restricted the use of cross validation. Additionally, all Whole Slide Images (WSI) were scanned with the same device and only one person annotated the images. 
The practical annotation effort depends on a combination of boundary accuracy and the number of superpixels, but the importance of each metric can not be quantified yet. These aspects will be further explored in future work to evaluate the downstream utility for reducing annotation time with the proposed methods. 

\begin{acknowledgement}
This project is supported by the Bavarian State Ministry of Health and Care, project grants No. PBN-MGP-2010-0004-DigiOnko and PBN-MGP-2008-0003-DigiOnko. We also gratefully acknowledge the support from the Interdisciplinary Center for Clinical Research (IZKF, Clinician Scientist Program) of the Medical Faculty FAU Erlangen-Nürnberg.
\end{acknowledgement}

% This command generates the bibliography using the entries of the .bib file.
% Remove it only if you do not use a bibliography. 
\printbibliography

\end{document}